\documentclass[letterpaper]{article} % DO NOT CHANGE THIS
\usepackage{aaai25}  % DO NOT CHANGE THIS
\usepackage{times}  % DO NOT CHANGE THIS
\usepackage{helvet}  % DO NOT CHANGE THIS
\usepackage{courier}  % DO NOT CHANGE THIS
\usepackage[hyphens]{url}  % DO NOT CHANGE THIS
\usepackage{graphicx} % DO NOT CHANGE THIS
\usepackage{natbib}  % DO NOT CHANGE THIS AND DO NOT ADD ANY OPTIONS TO IT
\usepackage{caption} % DO NOT CHANGE THIS AND DO NOT ADD ANY OPTIONS TO IT
\usepackage{algorithm}
\usepackage{pifont}
\usepackage{booktabs}
\usepackage{tabularx}
\usepackage{multirow}
\usepackage{listings}
\usepackage{threeparttable}
\usepackage{amsmath}
\usepackage{algpseudocode}
\usepackage{cleveref}
\crefname{section}{§}{§§}
\usepackage{xcolor}
\usepackage{tcolorbox}
\definecolor{verylightgray}{rgb}{.97,.97,.97}
\usepackage{cleveref}

\usepackage{newfloat}
\usepackage{listings}
\DeclareCaptionStyle{ruled}{labelfont=normalfont,labelsep=colon,strut=off} % DO NOT CHANGE THIS
\floatstyle{ruled}
\newfloat{listing}{tb}{lst}{}
\floatname{listing}{Listing}
\title{SCALM: Detecting Bad Practices in Smart Contracts Through LLMs}
\author{
    Zongwei Li,
    Xiaoqi Li\thanks{Corresponding author: Xiaoqi Li},
    Wenkai Li,
    Xin Wang
}
\affiliations{
    School of Cyberspace Security, Hainan University, Haikou, 570228, China\\
    lizw1017@gmail.com, csxqli@ieee.org, liwenkai871@gmail.com, wxin98767@gmail.com
}

\usepackage{bibentry}
\begin{document}

\maketitle

\begin{abstract}
As the Ethereum platform continues to mature and gain widespread usage, it is crucial to maintain high standards of smart contract writing practices. While bad practices in smart contracts may not directly lead to security issues, they do elevate the risk of encountering problems. Therefore, to understand and avoid these bad practices, this paper introduces the first systematic study of bad practices in smart contracts, delving into over 35 specific issues. Specifically, we propose a large language models (LLMs)-based framework, \textsc{SCALM}. It combines Step-Back Prompting and Retrieval-Augmented Generation (RAG) to identify and address various bad practices effectively. Our extensive experiments using multiple LLMs and datasets have shown that \textsc{SCALM} outperforms existing tools in detecting bad practices in smart contracts.
\end{abstract}

% Uncomment the following to link to your code, datasets, an extended version or similar.
%
% \begin{links}
%     \link{Code}{https://aaai.org/example/code}
%     \link{Datasets}{https://aaai.org/example/datasets}
%     \link{Extended version}{https://aaai.org/example/extended-version}
% \end{links}
\section{Introduction}
With the widespread use of blockchain technology, smart contracts have become an important part of the blockchain ecosystem \cite{Sharma_2023_review}. Smart contracts are computer programs that automatically execute contract terms, controlling assets and operations on the chain. However, due to their public and immutable code, smart contracts have become a significant target for attackers \cite{li2023overview}. A total of 464 security incidents occurred in 2023, resulting in losses of up to \$2.486 billion \cite{slowmist}. 
% The most significant attack occurred on September 23rd when Mixin Network's cloud service provider database was attacked, involving approximately \$200 million.

\textbf{Bad practices} refer to poor coding habits or design decisions in the development of smart contracts. Although bad practices may not result in immediate security threats, they could potentially lead to future issues such as performance problems, increased security vulnerabilities, and unpredictable code behavior \cite{kong2024characterizing}. Additionally, bad practices have hidden economic dangers due to the disruption of regular smart contract activities \cite{niu2024unveiling}.
 
Currently, the security audit of smart contracts mainly relies on manual code review and automated tools. However, these methods have their limitations \cite{{Chaliasos_2024_Smart}}. Manual code review is inefficient and prone to overlook subtle security vulnerabilities. Existing automated tools primarily rely on pattern matching, which cannot accurately detect complex security issues \cite{li2024cobra,li2020characterizing}. 
Moreover, the types of vulnerabilities that these tools can detect are usually relatively limited.
% and may not be able to identify all potential security problems in smart contracts. 
To achieve a comprehensive audit, multiple tools may be required, each covering different aspects of security. 
% Therefore, effectively detecting and preventing security issues in smart contracts remains an important issue that needs to be solved.

To solve these problems, we introduce \textsc{SCALM} (\textbf{S}mart \textbf{C}ontract \textbf{A}udit \textbf{L}anguage \textbf{M}odel), a new security auditing framework for smart contracts. It mainly consists of two parts: first, it performs static analysis on large datasets to identify and extract code blocks containing potential bad practices. These are then vectorized and stored in a vector database as a searchable knowledge base. Second, \textsc{SCALM} utilizes RAG and Step-Back Prompting to abstract high-level concepts and principles from the code, enabling the detection of bad practices. The framework ultimately generates detailed audit reports that highlight security issues, assess associated risks, and provide actionable remediation recommendations.

% The main contributions of this paper are as follows:
Our contributions are as follows:
\begin{itemize}
\item To the best of our knowledge, we provide the first systematic study of bad practices in smart contracts and conduct an in-depth discussion and analysis on 35 different types of SWC covered.

\item We introduce \textsc{SCALM}, a novel framework based on LLMs for smart contract security audits. It mainly consists of two parts: firstly, static code analysis is used to extract code blocks containing bad practices, which are transformed into vectors and stored in a vector database as a knowledge base; secondly, using RAG and Step-Back Prompting to extract high-level concepts and principles from the code to detect bad practices, and ultimately generate detailed audit reports.

\item We conduct an experimental evaluation on \textsc{SCALM}, and the results show that the framework performs well and outperforms existing tools in detecting bad practices in smart contracts. At the same time, ablation experiments reveal that the RAG component significantly improves \textsc{SCALM} performance.
% At the same time, we also found through ablation experiments that the RAG component has played a significant role in improving performance in the \textsc{SCALM}.

\item We open source \textsc{SCALM}'s codes and experimental data at \url{https://figshare.com/s/5cc3639706e4ecd16724}.
\end{itemize}

\section{Background}
\subsection{Large Language Models}
LLMs are trained using deep learning techniques to understand and generate human language.  They are typically based on the Transformer architecture, such as ChatGPT \cite{Kasneci_2023_ChatGPT}, BERT \cite{Devlin_2019_BERT}, GLM \cite{Du_2022_GLM}, etc.
The training process of these models usually involves learning language patterns and structures from large-scale text corpora. These corpora can include a variety of texts such as news articles, books, web pages, and other forms of human linguistic expression. The model can generate coherent and meaningful text by learning from these corpora. 
In addition, they can handle various natural language processing tasks, including text generation, text classification, sentiment analysis, question-answering systems, etc \cite{mao2024scla}.

One of the key features of LLMs is their powerful generative ability. These models can generate new, coherent text similar in grammar and semantics to the training data. This makes LLMs useful for various applications, including machine translation, text summarization, sentiment analysis, dialogue systems, and other natural language processing tasks.

% Another important feature 

Another important feature of LLMs is their "zero-shot" capability, which allows them to perform various tasks without any task-specific training. For example, the model can choose the most appropriate answer given a question and some answer options. This ability makes LLMs very useful in many practical applications.

However, LLMs also have some challenges and limitations. For instance, they may generate inaccurate or misleading information and reflect biases in the training data. Moreover, due to their need for substantial computational resources for training and operation, they might face certain challenges in practical application.

\subsection{Smart Contract Weakness Classification}  

Smart contracts are self-executing protocols that run on the blockchain and allow trusted transactions without third-party intervention. However, since their code is publicly available and cannot be changed once deployed, the security of smart contracts has become an important issue. To address this issue, EIP-1470 \cite{wagner2018eip} proposes the Smart Contract Weakness Classification (SWC), a tool designed to help developers identify and prevent smart contract weaknesses.

SWC concerns weaknesses that can be identified within a smart contract's Solidity code. It is designed to reference the structure and terminology of the Common Weakness Enumeration (CWE) but adds several weakness classifications specific to smart contracts. These classifications include but are not limited to, reentry attacks, arithmetic overflow, Assert Violation, etc.

All work on SWC has been incorporated into the EEA EthTrust Security Level Specification, a specification proposed by the Enterprise Ethereum Alliance (EEA) to provide a reliable methodology for assessing the security of smart contracts. 
This specification defines a series of security levels to measure the security and trustworthiness of smart contracts.

\section{Method}
\begin{figure}[tbp]
\centering
\includegraphics[width=\linewidth]{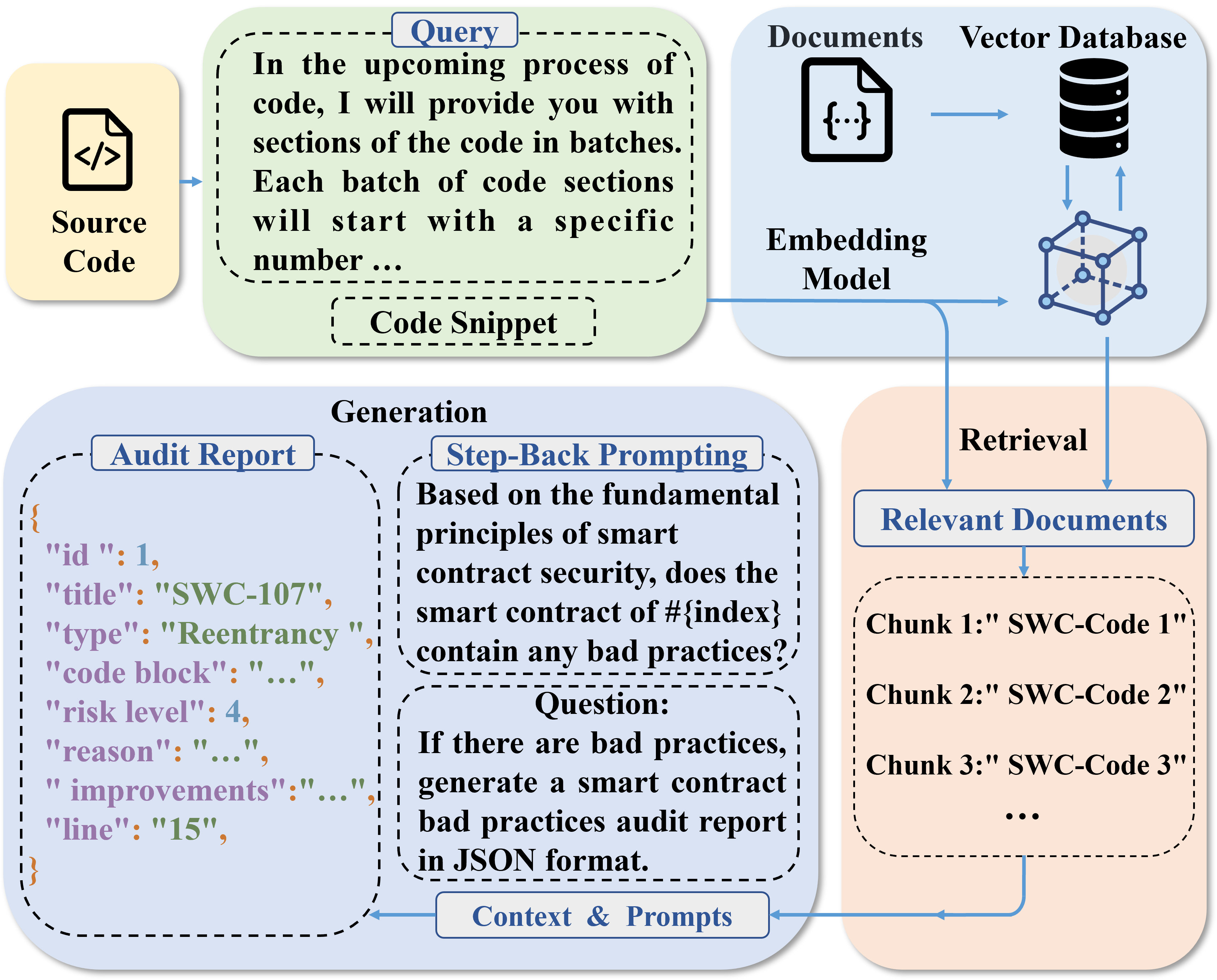}
\caption{Overall Architecture of SCALM.}
\label{fig:llm}
\vspace{-2em}
\end{figure}

The overview of \textsc{SCALM}'s architecture is shown in Fig.~\ref{fig:llm}, which mainly consists of two modules:
\underline{(1)} Extracts defective code blocks via static analysis, converts them into vectors, and stores them in a vector database to create a queryable library of code information.
\underline {(2)} Adopts the RAG methodology, which utilizes Step-Back Prompting to abstract high-level concepts and principles from the code and generates a detailed audit report.

The  Algorithm  \ref{alg:scalm} outlines the step-by-step procedure for generating an audit report for smart contract code. The algorithm is divided into several stages: initialization, data collection and processing, detection strategy, and report generation. 

\begin{algorithm}[tb]
\caption{SCALM Algorithm}
\label{alg:scalm}
\begin{algorithmic}[1]

\State \textbf{Input:} Smart contract code $C$, DAppSCAN database $D$
\State \textbf{Output:} Audit report $R$ in \texttt{JSON} format

\State \textbf{Initialization:}
\State Initialize vector database $VDB$
\State Initialize embedding model $f_{\text{Embedding}}$
\State Initialize LLM with RAG and Step-Back Prompting capabilities

\State \textbf{Data Collection and Processing:}
\For{each contract $c \in D$}
    \For{each code snippet $s \in c$}
        \If{$s$ contains bad practices}
            \State Convert $s$ to vector $\vec{v}_s$ using $f_{\text{Embedding}}$
            \State Store $\vec{v}_s$ in $VDB$
        \EndIf
    \EndFor
\EndFor

\State \textbf{Detection Strategy and Report Generation:}
\State Split $C$ into code fragments $F$
\State Initialize empty report $R$

\For{each fragment $f \in F$}
    \State Convert $f$ to vector $\vec{v}_f$ using $f_{\text{Embedding}}$
    \State Retrieve similar vectors $\{\vec{v}_i\}$ from $VDB$ using cosine similarity
    \State Retrieve corresponding code snippets $\{s_i\}$ for $\{\vec{v}_i\}$
    
    \State Perform Step-Back Prompting:
    \State Abstract higher-level concepts or principles from $\{s_i\}$
    \State Generate reasoning and evaluation using LLM
    
    \State Generate audit results for $f$:
    \State \textit{bad practice ID}, \textit{title}, \textit{type}, \textit{bad code block}, \textit{location}, \textit{risk level}, \textit{reason}, \textit{suggestions}, \textit{line}
    
    \State Append audit results to $R$
\EndFor

\State \textbf{Return} $R$

\end{algorithmic}
\end{algorithm}

\subsection{Data Collection and Processing}\label{dataset}
Our data collection comes from the DAppSCAN database \cite{zheng2023dappscan}, which includes 39,904 smart contracts with 1,618 SWC weaknesses. First, we extract code snippets with bad practices from these contracts. This extraction process involves static code analysis, using tools to identify and extract code snippets with potential security risks. Then, each code block is converted into a vector through our embedding model and stored in the vector database for subsequent fast matching and retrieval operations.

\subsection{Vector Database}
% We use a vector database to store and query large amounts of vector data. 
A vector database is a particular database that can store and query large amounts of vector data \cite{Hambardzumyan_2022_Deepa}. In the vector database, data is stored as vectors, each typically represented by a set of floating-point numbers. These vectors can represent various data types, such as images, audio, text, etc. In bad practice detection tasks, we use an embedding model (i.e., text-embedding-ada-002) to convert code into vectors and then store these vectors in the vector database \cite{li2024stateguard, li2024guardians}. This process can be expressed with the following \cref{eqa:embedding}:

\begin{equation}
\begin{aligned}
    \vec{v} = f_{\text{Embedding}}(\text{Text})
\label{eqa:embedding}
\end{aligned}
\end{equation}

Where $f_{\text{Embedding}}$ is our embedding model, $\text{Text}$ is the input text, and $\vec{v}$ is the outputted vector.
This vectorized data storage method significantly improves efficiency in handling it. Firstly, storing data as vectors makes it more compact, thus reducing storage space requirements. Secondly, vectorized data facilitates parallel computing, which is crucial when dealing with large-scale datasets. A vital feature of a Vector Database lies in its ability to perform efficient similarity searches, which are notably advantageous when dealing with high-dimensional datasets. This similarity search can be achieved by calculating cosine similarities between two vectors  with the \cref{eqa:similar}:

\begin{equation}
\begin{aligned}
    \text{similarity}(A,B)= \frac {A \cdot B}{||A||_2 \cdot ||B||_2}
\label{eqa:similar}
\end{aligned}
\end{equation}

where $A$ and $B$ are two vectors, $A \cdot B$ is their dot product, and $||A|||_2$ and $||B|||_2$ are their second-paradigms.
% second-paradigms (i.e., lengths)

During the query process, the system first converts user queries into vectors and then performs efficient matching and retrieval in the vector database. In the final stage of the query, the system converts matched results back into a format that users can understand and use. 

\subsection{Retrieval-Augmented Generation}

LLMs have proven their powerful capabilities in handling complex language understanding and generation tasks, incredibly when fine-tuned for specific downstream tasks. 
However, these models still face challenges for tasks that require precise and specific knowledge, such as smart contract code auditing.

\textsc{SCALM} adopts the RAG fine-tuning method to improve the quality and accuracy of smart contract code auditing. This method combines pre-trained parametric models with non-parametric memory to enhance the quality and accuracy of smart contract code audits \cite{Gao_2024_Retrieval-Augmented}. The RAG integrates the processes of retrieval and generation into one. 

\begin{figure}[tbp]
\centering
\includegraphics[width=\linewidth]{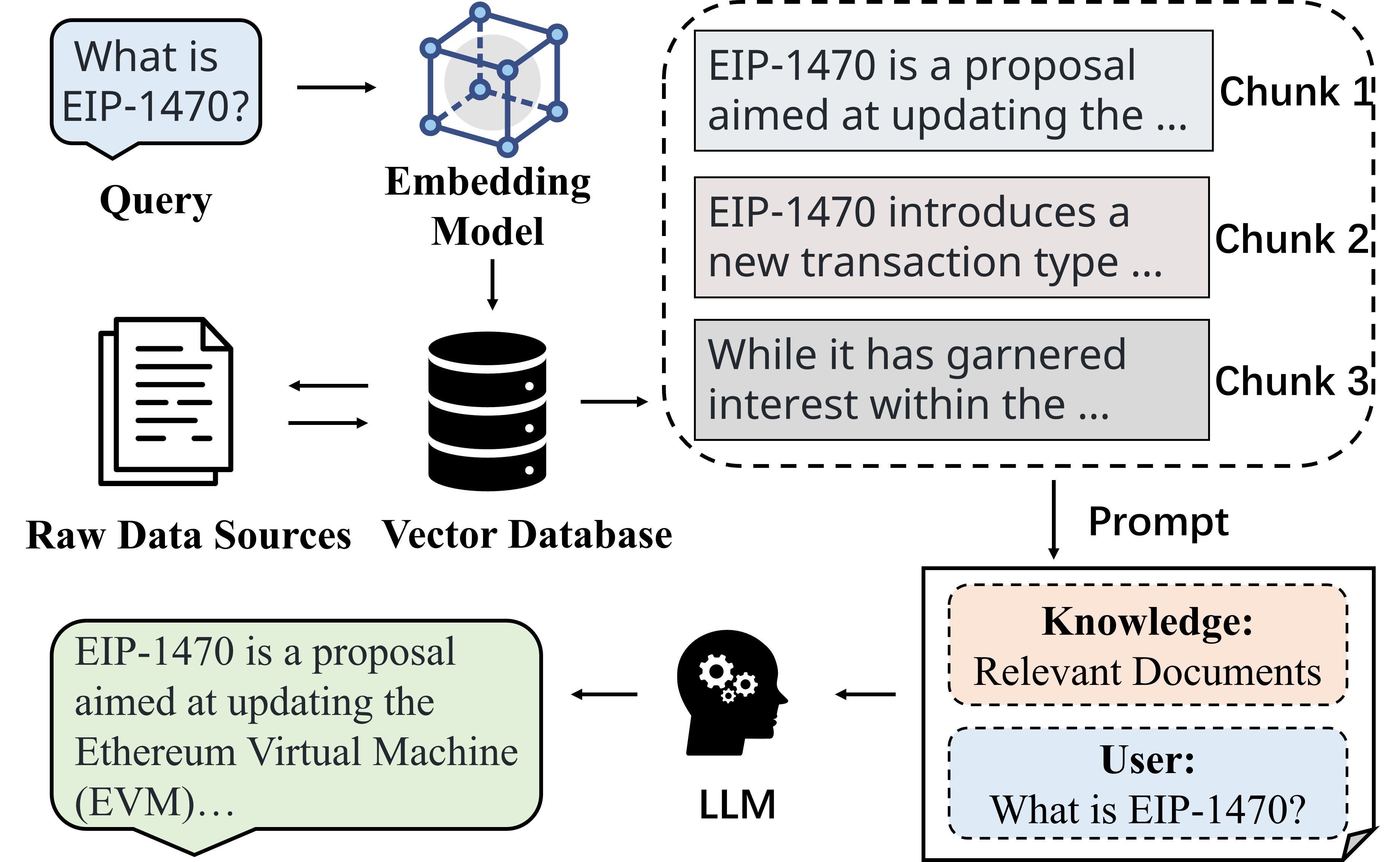}
\caption{Illustration of the RAG.}
\label{fig:rag}
\end{figure}

As Fig. \ref{fig:rag} illustrates, during the operation of the model, it first retrieves relevant documents or entities from a large-scale knowledge base. Then, it inputs this retrieved information as additional context into the generation model, which generates corresponding outputs based on these inputs. This design allows RAG to utilize external knowledge bases effectively while demonstrating excellent performance when dealing with tasks requiring extensive background knowledge.

\subsection{Detection Strategy and Report Generation}
Our detection strategy mainly focuses on effectively identifying bad practices in smart contracts. Firstly, the smart contract code to be detected is cut into smaller fragments according to its structure and size, then queried through a vector database to find the most matching known bad practice code blocks.

In the task of detecting bad practices in smart contracts, we adopted a technique called Step-Back Prompting \cite{Zheng__TAKE}. This technique leverages the capabilities of LLMs to abstract high-level concepts and basic principles from specific code instances. In this way, not only can the model understand the literal meaning of the code, but it can also comprehend underlying logic and potential design patterns through abstract thinking.

% Step-Back Prompting consists of the following two main steps:
Step-Back Prompting consists of two main steps:
\begin{itemize} 
\item Abstraction: Instead of directly posing questions, we propose a general step-back question about higher-level concepts or principles and retrieve facts related to these high-level concepts or principles. In the task of detecting bad practices in smart contracts, we use abstract prompts that aim to guide LLMs to explore not just the literal meaning of code but deeper structures and intentions. These prompts may include questions like "What are the potential risks with this implementation?" 

% or "Does this method comply with basic principles for secure smart contracts?"

\item Model Reasoning: Based on facts about high-level concepts or principles, LLMs can reason about answers to the original question. We refer to this as abstraction-based reasoning. Reasoning with these abstraction hints attempts to analyze the code from a broader perspective. This includes comparing the strengths and weaknesses of different implementations and how they fit with known best practices or common bad practices.
\end{itemize}

The LLM evaluates these code snippets for bad practices through Step-Back Prompting and generates a detailed audit report based on the results. The report is output in \texttt{JSON} format. It contains the  \textit{bad practice ID},  \textit{title}, \textit{type}, specific \textit{bad code block} along with its \textit{location}, \textit{risk level},  \textit{reason} for the problem, and \textit{suggestions} for improvement.

Through this method, we can effectively utilize the powerful capabilities of LLMs for deep security audits on smart contracts, thereby helping developers identify and fix potential security issues.

\section{Experiments}
\subsection{Experiment Settings}
All experiments are executed on a server equipped with NVIDIA GeForce GTX 4070Ti GPU, Intel(R) Core(TM) i9-13900KF CPU, and 128G RAM, operating on Ubuntu 22.04 LTS. The software environment includes Python 3.9 and PyTorch 2.0.1.

\noindent\textbf{Dataset.  }
In this paper, we use the \textbf{DAppSCAN} dataset \cite{zheng2023dappscan} as a knowledge base for detecting bad practices in smart contracts. The dataset contains 39,904 Solidity files with 1,618 SWC weaknesses from 682 real projects. The \textbf{Smartbugs} dataset \cite{DurieuxEtAl2020ICSE} is also used in the experiment, and a total of 1,894 smart contracts with five types of SWC weaknesses are extracted for comparison experiments. These datasets form the basis of our experimental analysis. Table \ref{table:dataset} summarizes the smart contract data used.

We conduct initial detection experiments using official SWC samples covering 35 categories of bad practices. It is important to note that these 35 categories of SWC include most bad practices, but they do not represent all possible bad practices that may occur in smart contract development. In the follow-up experiments, we select five categories of SWC for in-depth study, using 94 samples for positive examples of SWC-104 and 200 samples for both positive and negative examples of the remaining SWC categories. In order to comprehensively evaluate the performance of the model on the test set, we select Accuracy (Acc), Recall, and F1 score as the evaluation metrics.

\noindent\textbf{Models.  }
For the selection of LLMs, we chose six current state-of-the-art models for detection experiments. GPT-4o, GPT-4-1106-preview, and GPT-4-0409 are the latest versions from OpenAI with powerful natural language processing capabilities; Claude-3.5-Sonnet is Anthropic's new-generation model focused on safety and interpretability; Gemini-1.5-Pro is Google's high-performance model optimized for multitasking; Llama-3.1-70b-Instruct is Meta's large-scale model specializing in instruction following and generating high-quality text.

\begin{table}[tb]
\centering
\small
\begin{tabular}{>{\centering\arraybackslash}p{2.2cm} c c}
\toprule
\textbf{Dataset} & \textbf{\# Contracts}  \\
\midrule
DAppSCAN & 39,904  \\
\midrule
Smartbugs & 1,894  \\
\bottomrule
\end{tabular}
\caption{The Collected Dataset for Our Evaluation. \# indicates the number of each item.}
\label{table:dataset}
\vspace{-2em}
\end{table}

\begin{table*}[tb]
\centering
\resizebox{\textwidth}{!}{% Resize table to fit within column width
\begin{tabular}{@{}lccccccc@{}}
\toprule
\multirow{2}{*}{\textbf{SWC-ID}} & \multirow{2}{*}{\textbf{Title}} & \multicolumn{6}{c}{\textbf{Models}} \\
\cline{3-8}
 & & GPT-4o & GPT-4-1106 & GPT-4-0409 & Claude & Gemini & Llama \\
\midrule
SWC-100 & Function Default Visibility & \ding{51} & \ding{55} & \ding{55} & \ding{55} & \ding{55} & \ding{55} \\
SWC-101 & Integer Overflow and Underflow & \ding{51} & \ding{51} & \ding{51} & \ding{51} & \ding{51} & \ding{51} \\
SWC-102 & Outdated Compiler Version & \ding{51} & \ding{51} & \ding{51} & \ding{51} & \ding{51} & \ding{51} \\
SWC-103 & Floating Pragma & \ding{51} & \ding{51} & \ding{51} & \ding{51} & \ding{51} & \ding{51} \\
SWC-104 & Unchecked Call Return Value & \ding{51} & \ding{51} & \ding{51} & \ding{51} & \ding{51} & \ding{51} \\
SWC-105 & Unprotected Ether Withdrawal & \ding{51} & \ding{51} & \ding{51} & \ding{51} & \ding{51} & \ding{51} \\
SWC-106 & Unprotected SELFDESTRUCT Instruction & \ding{51} & \ding{51} & \ding{51} & \ding{51} & \ding{51} & \ding{51} \\
SWC-107 & Reentrancy & \ding{51} & \ding{51} & \ding{51} & \ding{51} & \ding{51} & \ding{51} \\
SWC-108 & State Variable Default Visibility & \ding{51} & \ding{51} & \ding{51} & \ding{51} & \ding{51} & \ding{51} \\
SWC-109 & Uninitialized Storage Pointer & \ding{55} & \ding{55} & \ding{55} & \ding{55} & \ding{55} & \ding{55} \\
SWC-110 & Assert Violation & \ding{51} & \ding{51} & \ding{51} & \ding{51} & \ding{51} & \ding{51} \\
SWC-111 & Use of Deprecated Solidity Functions & \ding{51} & \ding{51} & \ding{51} & \ding{51} & \ding{51} & \ding{51} \\
SWC-112 & Delegatecall to Untrusted Callee & \ding{51} & \ding{51} & \ding{51} & \ding{51} & \ding{55} & \ding{55} \\
SWC-113 & DoS with Failed Call & \ding{51} & \ding{51} & \ding{51} & \ding{51} & \ding{55} & \ding{51} \\
SWC-114 & Transaction Order Dependence & \ding{55} & \ding{55} & \ding{55} & \ding{55} & \ding{51} & \ding{55} \\
SWC-115 & Authorization through tx.origin & \ding{51} & \ding{51} & \ding{51} & \ding{51} & \ding{51} & \ding{51} \\
SWC-116 & Block values as a proxy for time & \ding{51} & \ding{51} & \ding{51} & \ding{51} & \ding{51} & \ding{51} \\
SWC-117 & Signature Malleability & \ding{51} & \ding{51} & \ding{51} & \ding{55} & \ding{55} & \ding{51} \\
SWC-118 & Incorrect Constructor Name & \ding{51} & \ding{55} & \ding{51} & \ding{51} & \ding{55} & \ding{55} \\
SWC-119 & Shadowing State Variables & \ding{51} & \ding{51} & \ding{51} & \ding{51} & \ding{51} & \ding{51} \\
SWC-120 & Weak Sources of Randomness from Chain Attributes & \ding{51} & \ding{51} & \ding{51} & \ding{51} & \ding{51} & \ding{51} \\
SWC-123 & Requirement Violation & \ding{55} & \ding{51} & \ding{55} & \ding{51} & \ding{51} & \ding{51} \\
SWC-124 & Write to Arbitrary Storage Location & \ding{51} & \ding{55} & \ding{51} & \ding{55} & \ding{55} & \ding{55} \\
SWC-125 & Incorrect Inheritance Order & \ding{55} & \ding{55} & \ding{55} & \ding{55} & \ding{55} & \ding{55} \\
SWC-126 & Insufficient Gas Griefing & \ding{51} & \ding{51} & \ding{51} & \ding{55} & \ding{55} & \ding{55} \\
SWC-127 & Arbitrary Jump with Function Type Variable & \ding{51} & \ding{51} & \ding{51} & \ding{51} & \ding{51} & \ding{51} \\
SWC-128 & DoS With Block Gas Limit & \ding{51} & \ding{51} & \ding{51} & \ding{51} & \ding{51} & \ding{51} \\
SWC-129 & Typographical Error & \ding{51} & \ding{51} & \ding{51} & \ding{51} & \ding{51} & \ding{51} \\
SWC-130 & Right-To-Left-Override control character (U+202E) & \ding{55} & \ding{55} & \ding{55} & \ding{55} & \ding{51} & \ding{51} \\
SWC-131 & Presence of unused variables & \ding{51} & \ding{51} & \ding{51} & \ding{55} & \ding{51} & \ding{55} \\
SWC-132 & Unexpected Ether balance & \ding{51} & \ding{51} & \ding{51} & \ding{51} & \ding{55} & \ding{55} \\
SWC-133 & Hash Collisions With Multiple Variable Length Arguments & \ding{55} & \ding{55} & \ding{55} & \ding{55} & \ding{55} & \ding{55} \\
SWC-134 & Message call with hardcoded gas amount & \ding{51} & \ding{51} & \ding{51} & \ding{51} & \ding{51} & \ding{51} \\
SWC-135 & Code With No Effects & \ding{51} & \ding{51} & \ding{51} & \ding{51} & \ding{51} & \ding{51} \\
SWC-136 & Unencrypted Private Data On-Chain & \ding{55} & \ding{55} & \ding{55} & \ding{51} & \ding{55} & \ding{55} \\
\bottomrule
\end{tabular}
}
\caption{SWC bad practice detection. Full model names are GPT-4o, GPT-4-1106-preview, GPT-4-0409, Claude-3.5-Sonnet, Gemini-1.5-Pro, and Llama-3.1-70b-Instruct. A checkmark (\ding{51}) indicates successful detection and a cross (\ding{55}) indicates a failure.}
\label{table:swc}
\vspace{-1em}
\end{table*}

\noindent\textbf{Evaluation Metrics.   } We carry out experiments to answer the following research questions:
\underline{\textbf{RQ1:}} How effective is \textsc{SCALM} in detecting bad practices in smart contracts? How do different LLMs affect \textsc{SCALM}?
\underline{\textbf{RQ2:}}  Can \textsc{SCALM} find bad practices undetectable by other tools? How does it compare with existing tools?
\underline{\textbf{RQ3:}} Can \textsc{SCALM} achieve the same effect without including RAG?

\subsection{RQ1: Bad Practice Detection}
To assess the effectiveness of \textsc{SCALM} in detecting bad practices within smart contracts, we conduct a comprehensive evaluation using a variety of LLMs. The primary objective is to determine how well \textsc{SCALM} identifies known bad practices, as classified by the Smart SWC registry. We also seek to understand the impact of different LLMs on the performance of \textsc{SCALM} in detecting these bad practices.

We use the DAppSCAN dataset containing 23637 smart contracts as the knowledge base to test the detection ability of \textsc{SCALM}. Specifically, we focus on contracts with known SWC vulnerabilities. Our experiments involve running \textsc{SCALM} with multiple LLM configurations, including GPT-4o, GPT-4-1106-preview, GPT-4-0409, Claude-3.5-Sonnet, Gemini-1.5-Pro, and Llama-3.1-70b-Instruct.

For each model, we evaluate the framework's ability to identify instances of bad practices across 35 SWC categories correctly. Each detection instance is recorded as a success or failure based on whether the LLM accurately identifies the bad practice by its SWC-ID or keywords.

The results of the experiments are summarized in Table~\ref{table:swc}. The table shows the detection capabilities of each LLM across the different SWC categories, with a checkmark (\ding{51}) indicating successful detection and a cross (\ding{55}) indicating a failure.

Among the LLMs evaluated, GPT-4o demonstrates the highest detection accuracy, successfully identifying bad practices across 33 of the 36 SWC categories. This model is particularly effective in detecting vulnerabilities such as \emph{Integer Overflow and Underflow} (SWC-101), \emph{Outdated Compiler Version} (SWC-102), and \emph{Reentrancy} (SWC-107).

However, the performance varies significantly among the different LLMs. For instance, while models like GPT-4-1106-preview and GPT-4-0409 also perform well, they exhibit some inconsistencies, such as missing detections in categories like \emph{Function Default Visibility} (SWC-100).

On the other hand, LLMs like Claude-3.5-Sonnet, Gemini-1.5-Pro, and Llama-3.1-70b-Instruct have a more mixed performance, with successful detections in specific categories but noticeable gaps in others. For example, Gemini-1.5-Pro fails to detect \emph{Delegatecall to Untrusted Callee} (SWC-112) and \emph{Transaction Order Dependence} (SWC-114), while Claude-3.5-Sonnet misses issues related to \emph{Signature Malleability} (SWC-117) and \emph{Incorrect Constructor Name} (SWC-118).

The variations in detection accuracy across different LLMs highlight the importance of model selection in the \textsc{SCALM} framework. The results suggest that while models like GPT-4o can handle a wide range of bad practices effectively, others may require further fine-tuning or additional contextual information to improve their performance.

Additionally, the findings emphasize the need for a robust and diverse LLM ensemble within \textsc{SCALM} to ensure comprehensive coverage of all potential bad practices in smart contracts. By leveraging the strengths of different models, \textsc{SCALM} can achieve a more reliable and thorough detection process, ultimately leading to higher-quality audit reports.

\begin{tcolorbox}[boxrule=1pt,boxsep=1pt,left=2pt,right=2pt,top=2pt,bottom=2pt]
\textbf{Answer to RQ1.}
\textsc{SCALM} is a powerful framework for detecting bad practices in smart contracts, especially when supported by advanced LLMs like GPT-4o. However, the choice of LLM significantly impacts the system's overall effectiveness, indicating that ongoing model improvements and updates are essential for maintaining high detection accuracy.
\end{tcolorbox}

\subsection{RQ2: Comparison Experiments}
We conduct a series of comparison experiments to address RQ2, which examines whether \textsc{SCALM} can identify bad practices that other tools cannot detect and how it compares with existing tools. These experiments focus on several critical SWC categories, specifically SWC-101 (\emph{Integer Overflow and Underflow}), SWC-104 (\emph{Unchecked Call Return Value}), SWC-107 (\emph{Reentrancy}), SWC-112 (\emph{Delegatecall to Untrusted Callee}), and SWC-116 (\emph{Block Values as a Proxy for Time}). We select 1,894 smart contracts from the SmartBugs dataset containing these five categories of bad practices. 
% We conduct a series of comparison experiments to address RQ2, which examines whether \textsc{SCALM} can identify bad practices that other tools cannot detect and how it compares with existing tools. These experiments focus on several critical SWC categories, specifically SWC-101, SWC-104, SWC-107, SWC-112, and SWC-116. We select 1,894 smart contracts from the SmartBugs dataset containing these five categories of bad practices. 

Additionally, we collect a set of smart contract defect detection tools from reputable journals and conferences in software and security (e.g., CCS and ASE) as well as Mythril \cite{mythril}, recommended by the official Ethereum community.
For comparative analysis, we choose four benchmark smart contract detection tools, namely Mythril, Oyente \cite{Luu_2016_Making}, Confuzzius \cite{Torres_2021_ConFuzzius}, and Conkas \cite{VelosoConkas}. Several factors are considered in the selection of the tools: \underline{(1)} the accessibility of the tool's source code; \underline{(2)} the tool's ability to detect the five categories of bad practices we select; \underline{(3)} the tool's support for source code written in Solidity; \underline{(4)} the tool's ability to report the exact location of potentially defective code for manual review. 
% To ensure the fairness of the experimental results, we standardize the test environment and parameter configuration of all tools.

\begin{table}[tb]
\centering
\small
\begin{threeparttable}
\begin{tabular}{lccccc}
\toprule
\textbf{SWC-ID} & \textbf{Tools} & \textbf{Acc(\%)} & \textbf{Recall(\%)} & \textbf{F1(\%)} \\
\midrule
\multirow{5}{*}{SWC-101} & conkas & 49.27 & 67.20 & 59.10 \\
 & mythril & 48.03 & 16.67 & 25.70 \\
 & oyente & 62.33 & 69.35 & 66.03 \\
 & confuzzius & 50.26 & 10.55 & 18.03 \\
 & \textbf{SCALM} & \textbf{95.50} & \textbf{94.50} & \textbf{95.45} \\
\midrule
\multirow{5}{*}{SWC-104} & conkas & 59.88 & 20.12 & 33.00 \\
 & mythril & 56.89 & 16.28 & 28.00 \\
 & oyente & — & — & — \\
 & confuzzius & 52.32 & 12.17 & 20.81 \\
 & \textbf{SCALM} & \textbf{98.25} & \textbf{99.50} & \textbf{98.27} \\
\midrule
\multirow{5}{*}{SWC-107} & conkas & 71.79 & 94.51 & 77.50 \\
 & mythril & 68.12 & 62.94 & 69.26 \\
 & oyente & 59.12 & 14.04 & 24.49 \\
 & confuzzius & 40.00 & 1.05 & 2.04 \\
 & \textbf{SCALM} & \textbf{95.00} & \textbf{94.50} & \textbf{94.97} \\
\midrule
\multirow{5}{*}{SWC-112}  & conkas & — & — & — \\
 & mythril & 86.67 & 54.29 & 70.37 \\
 & oyente & — & — & — \\
 & confuzzius & 78.86 & 7.14 & 13.33 \\
 & \textbf{SCALM} & \textbf{98.30} & \textbf{95.74} & \textbf{97.30} \\

\midrule
\multirow{5}{*}{SWC-116} & conkas & 89.35 & 86.99 & 88.50 \\
 & mythril & 76.45 & 50.41 & 63.87 \\
 & oyente & 48.20 & 3.16 & 6.03 \\
 & confuzzius & — & — & — \\
 & \textbf{SCALM} & \textbf{93.00} & \textbf{88.50} & \textbf{92.67} \\
\bottomrule
\end{tabular}
\vspace{-1em}
\end{threeparttable}
\caption{Comparison Experiments. SWC-101 is Integer Overflow and Underflow, SWC-104 is Unchecked Call Return Value, SWC-107 is Reentrancy, SWC-112 is Delegatecall to Untrusted Callee, and SWC-116 uses Block Values as a Proxy for Time.}
\label{table:swc3}
\end{table}

In these experiments, \textsc{SCALM} is powered by GPT-4o, one of the most advanced LLMs. For each SWC category, we evaluate the performance of \textsc{SCALM} and the other tools based on three key metrics: Accuracy (Acc), Recall, and F1 Score. These metrics provide a comprehensive view of each tool's detection capabilities:
\begin{itemize}
    \item \textbf{Accuracy (Acc)} measures the proportion of correctly identified instances (both true positives and true negatives) out of the total instances.
    \item \textbf{Recall} measures the proportion of actual positive instances correctly identified by the tool.
    \item \textbf{F1 Score} is the harmonic mean of Precision and Recall, providing a single metric that balances false positives and false negatives.
\end{itemize}

The results of these experiments are summarized in Table~\ref{table:swc3}. The results indicate that \textsc{SCALM} outperforms other tools across all evaluated SWC categories. For instance, in SWC-101 (\emph{Integer Overflow and Underflow}), \textsc{SCALM} achieves an F1 score of 95.45\%, significantly higher than Oyente's 66.03\%. Similarly, in SWC-104 (\emph{Unchecked Call Return Value}), \textsc{SCALM} reaches an F1 score of 98.27\%, while Conkas has only 33.00\%. In SWC-107 (\emph{Reentrancy}), \textsc{SCALM} scores 94.97\%, compared to Conkas's 77.50\%. For SWC-112 (\emph{Delegatecall to Untrusted Callee}), \textsc{SCALM} achieves an F1 score of 97.30\%, whereas Mythril manages 70.37\%. Lastly, in SWC-116 (\emph{Block Values as a Proxy for Time}), \textsc{SCALM} has an F1 score of 92.67\%, outperforming Conkas's 88.50\%. These results highlight \textsc{SCALM} 's superior Accuracy and Recall in detecting a range of smart contract bad practices.

The comparison experiments demonstrate that \textsc{SCALM} effectively detects bad practices in smart contracts, outperforming existing tools across multiple SWC categories. 
This suggests that \textsc{SCALM} can provide a more comprehensive and accurate analysis of smart contract security.

\begin{tcolorbox}[boxrule=1pt,boxsep=1pt,left=2pt,right=2pt,top=2pt,bottom=2pt]
\textbf{Answer to RQ2.}
\textsc{SCALM} detects bad practices more accurately than existing tools and identifies bad practices that others often miss. 
Across the five SWC categories tested, \textsc{SCALM} consistently achieves higher Accuracy, Recall, and F1 scores.
\end{tcolorbox}

\subsection{RQ3: Ablation Experiments}
We conduct ablation experiments to address RQ3, investigating whether \textsc{SCALM} can achieve the same performance without including the RAG component. In these experiments, \textsc{SCALM} is also powered by GPT-4o. We compare \textsc{SCALM} with and without RAG across several SWC categories: SWC-101, SWC-104, SWC-107, SWC-112, and SWC-116. The results are summarized in Table~\ref{table:swc_rag}.

\begin{table}[tbp]
\centering
\small
\begin{threeparttable}
\begin{tabular}{lccccc}
\toprule
\textbf{SWC-ID}  & \textbf{Tools} & \textbf{Acc(\%)} & \textbf{Recall(\%)}  & \textbf{F1(\%)} \\
\midrule
\multirow{2}{*}{SWC-101}  & w/o RAG & 79.25 & 81.50 & 79.71 \\
 & \textbf{SCALM} & \textbf{95.50} & \textbf{94.50}  & \textbf{95.45} \\
\midrule
\multirow{2}{*}{SWC-104}  & w/o RAG & 77.00 & 87.50  & 79.19 \\
 & \textbf{SCALM} & \textbf{98.25} & \textbf{99.50}  & \textbf{98.27} \\
\midrule
\multirow{2}{*}{SWC-107}  & w/o RAG & 73.50 & 81.00  & 75.35 \\
 & \textbf{SCALM} & \textbf{95.00} & \textbf{94.50}  & \textbf{94.97} \\
\midrule
\multirow{2}{*}{SWC-112}  & w/o RAG & 84.69 & 68.09  & 73.99 \\
 & \textbf{SCALM} & \textbf{98.30} & \textbf{95.74}  & \textbf{97.30} \\
\midrule
\multirow{2}{*}{SWC-116}  & w/o RAG & 82.75 & 85.00 & 83.13 \\
 & \textbf{SCALM} & \textbf{93.00} & \textbf{88.50}  & \textbf{92.67} \\
\bottomrule
\end{tabular}
\end{threeparttable}
\caption{Ablation Experiments. 'w/o RAG' indicates excluding the RAG component, while SCALM includes it.}
\label{table:swc_rag}
\vspace{-2em}
\end{table}

The ablation experiments reveal that including RAG significantly enhances \textsc{SCALM} 's performance across all SWC categories. For SWC-101 (\emph{Integer Overflow and Underflow}), \textsc{SCALM} with RAG achieves an F1 score of 95.45\%, compared to 79.71\% without RAG. In SWC-104 (\emph{Unchecked Call Return Value}), the F1 score with RAG is 98.27\%, while without RAG it is 79.19\%. Similarly, for SWC-107 (\emph{Reentrancy}), the F1 score increases from 75.35\% without RAG to 94.97\% with RAG. In SWC-112 (\emph{Delegatecall to Untrusted Callee}), \textsc{SCALM} with RAG achieves an F1 score of 97.30\%, compared to 73.99\% without RAG. Finally, for SWC-116 (\emph{Block Values as a Proxy for Time}), the F1 score with RAG is 92.67\%, whereas without RAG it is 83.13\%.
% The ablation experiments reveal that including RAG significantly enhances \textsc{SCALM} 's performance across all SWC categories. For SWC-101, \textsc{SCALM} with RAG achieves an F1 score of 95.45\%, compared to 79.71\% without RAG. In SWC-104, the F1 score with RAG is 98.27\%, while without RAG it is 79.19\%. Similarly, for SWC-107, the F1 score increases from 75.35\% without RAG to 94.97\% with RAG. In SWC-112, \textsc{SCALM} with RAG achieves an F1 score of 97.30\%, compared to 73.99\% without RAG. Finally, for SWC-116, the F1 score with RAG is 92.67\%, whereas without RAG it is 83.13\%.

The results demonstrate that the RAG component is crucial for \textsc{SCALM} 's high performance in detecting smart contract vulnerabilities. The significant drop in Accuracy, Recall, and F1 scores when RAG is excluded indicates that retrieving relevant knowledge and context provided by RAG substantially contributes to the model's effectiveness. Without RAG, \textsc{SCALM} struggles to achieve the same level of precision and Recall, underscoring the importance of this component in enhancing the model's overall capability to identify and report bad practices in smart contracts.

\begin{tcolorbox}[boxrule=1pt,boxsep=1pt,left=2pt,right=2pt,top=2pt,bottom=2pt]
\textbf{Answer to RQ3.}
The ablation experiments demonstrate that \textsc{SCALM} cannot achieve the same level of performance without the RAG component. Including RAG significantly improves F1 scores, Accuracy, and Recall across all tested SWC categories. 
\end{tcolorbox} 

\section{Discussion}
% The experimental results confirm \textsc{SCALM}'s effectiveness in detecting bad practices in smart contracts, outperforming existing tools in accuracy, recall, and F1 scores, particularly in identifying bad practices classified by the SWC registry. 
The experimental results confirm \textsc{SCALM}'s effectiveness in detecting bad practices in smart contracts, outperforming existing tools in accuracy, recall, and F1 scores.

The choice of LLM significantly impacts \textsc{SCALM}'s performance, with different models yield varying results. This finding emphasizes the need for careful model selection and ongoing improvements to maintain high accuracy. The RAG component is also crucial in \textsc{SCALM}. Ablation experiments showed significant performance drops when RAG was excluded. This highlights its importance in providing contextual information that enhances detection accuracy.

Despite its strengths, \textsc{SCALM} has limitations in understanding blockchain mechanisms and complex interactions. Specifically, \textsc{SCALM} might not fully grasp the state-dependent nature of certain vulnerabilities, which require a deep contextual understanding of how the contract evolves over time. 

Future work will focus on improving \textsc{SCALM} by incorporating advanced LLMs and fine-tuning with domain-specific data. Additionally, we plan to improve the RAG generation strategy to provide more prosperous and accurate contextual information. These improvements will enhance the performance of \textsc{SCALM} for smart contract auditing.

\section{Related Work}
Smart contract vulnerability detection has been extensively studied using various approaches. Traditional static analysis tools such as Slither \cite{Feist_2019_Slither}, Securify \cite{tsankov2018securify}, Ethainter \cite{brent2020ethainter}, Vandal \cite{brent2018vandal}, and eThor \cite{schneidewind2020ethor} employ pattern matching and dataflow analysis to identify vulnerabilities. Symbolic execution tools like Mythril \cite{mueller2018smashing}, Manticore \cite{Mossberg_2019_Manticore}, Osiris \cite{torres2018osiris}, and Pakala \cite{pakala} explore program paths to detect issues such as reentrancy and integer overflow. Fuzzing approaches including Echidna \cite{grieco2020echidna} and Harvey \cite{wustholz2020harvey} generate test inputs to uncover runtime vulnerabilities.

Recent work has explored deep learning methods for vulnerability detection. Graph neural network (GNN) based approaches \cite{Zhuang_2020_Smart, luo2024scvhunter, zhen2024gnn, 10700860, shang2025cegt} represent smart contracts as graphs to capture structural information. Cross-modality learning \cite{10.1145/3543507.3583367} and contrastive learning \cite{chen2024improving} have also been applied to enhance detection accuracy. Other studies focus on specific vulnerability types, such as reentrancy detection \cite{song2025silence}, state-inconsistency bugs \cite{liu2025detecting}, and NFT-specific defects \cite{ma2025uncovering}. Multimodal learning approaches like SmartInv \cite{10646885} combine multiple data representations for invariant inference. Deep learning with feature fusion \cite{LI2026122645} and bytecode analysis \cite{10.1145/3699597, wang2023smart} have also shown promising results.

LLMs have been widely applied and validated for their ability to identify and fix vulnerabilities \cite{Napoli_2023_Evaluating, chen2025chatgpt}. In the field of smart contract security, various methods based on LLMs have been proposed, and certain effects have been achieved.
Firstly, Boi et al. \cite{Boi_2024_VulnHunt-GPT} proposed VulnHunt-GPT, a method that uses the GPT-3 to identify common vulnerabilities in OWASP standards smart contracts.s. Similarly, Xia et al. \cite{Xia_2024_AuditGPT} introduced a tool called AuditGPT, which utilizes LLM to automatically and comprehensively verify the ERC rules of smart contracts. They break down large, complex audit processes into small, manageable tasks and design prompts for each ERC rule type to improve audit performance. 
In terms of fuzz testing, Shou et al. \cite{Shou_2024_LLM4Fuzz} presented LLM4FUZZ, which employs LLMs to effectively guide fuzz activities in smart contracts and determine their priorities. This approach can enhance test efficiency and coverage compared to traditional fuzz testing methods.

However, the direct use of pre-trained LLMs is no longer sufficient on some occasions, so many studies have chosen to fine-tune LLMs to meet specific needs. For example, Liu et al. \cite{Liu_2024_PropertyGPT} proposed PropertyGPT,  a system for formal verification of smart contracts by retrieving enhanced property generation. This system utilizes the capability of LLM to automatically generate comprehensive properties of smart contracts, including invariants, preconditions, and postconditions, thus improving the security of smart contracts. On the other hand, Storhaug et al. \cite{Storhaug_2023_Efficient} proposed a novel vulnerability-bound decoding method to reduce the amount of vulnerable code generated by the model. They fine-tuned the LLM to include vulnerability tags when generating code and then prohibited these tags during the decoding process to avoid producing vulnerable codes. Finally, Yang et al. \cite{Yang_2024_Automated} collected a large number of tagged smart contract vulnerabilities and fine-tuned Llama-2-13B for the automatic detection of vulnerabilities in the decentralized finance (DeFi) domain's smart contracts \cite{li2024defitail}.

In conclusion, while LLMs show great potential for improving smart contract security, the full realization of their potential requires continuous fine-tuning and adaptation of these models in specific contexts. Formal verification approaches \cite{runtimeverification} and dataset construction efforts \cite{10486822} also contribute to advancing the field.

\section{Conclusion}
This paper presents the first systematic study of over 35 bad practices in smart contracts. Building on this extensive analysis, we introduced \textsc{SCALM}, a framework based on LLMs for detecting these bad practices in smart contracts. Leveraging the power of LLMs, along with RAG and Step-Back Prompting, \textsc{SCALM} provides a comprehensive and accurate audit of smart contracts. Experiments on datasets such as DAppSCAN and SmartBugs show that \textsc{SCALM} outperforms existing tools, with the RAG component playing a critical role in improving detection accuracy. These findings demonstrate \textsc{SCALM}'s potential to enhance smart contract security, offering developers a robust framework to identify and address bad practices.

\section*{Acknowledgments}

This work is sponsored by the National Natural Science Foundation of China (No.62402146 \& 62362021).
% \section{Acknowledgments}
% AAAI is especially grateful to Peter Patel Schneider for his work in implementing the original aaai.sty file, liberally using the ideas of other style hackers, including Barbara Beeton. We also acknowledge with thanks the work of George Ferguson for his guide to using the style and BibTeX files --- which has been incorporated into this document --- and Hans Guesgen, who provided several timely modifications, as well as the many others who have, from time to time, sent in suggestions on improvements to the AAAI style. We are especially grateful to Francisco Cruz, Marc Pujol-Gonzalez, and Mico Loretan for the improvements to the Bib\TeX{} and \LaTeX{} files made in 2020.

\bibliography{aaai25}

@String{Computing = "Computing" }

@String{Computer = "{IEEE} Computer" }

@inproceedings{Boi_2024_VulnHunt-GPT,
  title = {{{VulnHunt-GPT}}: A {{Smart Contract}} Vulnerabilities Detector Based on {{OpenAI chatGPT}}},
  booktitle = {Proceedings of the {{ACM}}/{{SIGAPP Symposium}} on {{Applied Computing}} (SAC)},
  author = {Boi, Biagio and Esposito, Christian and Lee, Sokjoon},
  year = {2024},
  pages = {1517--1524}

}

@misc{Devlin_2019_BERT,
  title = {{{BERT}}: {{Pre-training}} of {{Deep Bidirectional Transformers}} for {{Language Understanding}}},
  author = {Devlin, Jacob and Chang, Ming-Wei and Lee, Kenton and Toutanova, Kristina},
  year = {2019},
  eprint = {1810.04805},
  archivePrefix={arXiv}
}

@misc{Du_2022_GLM,
  title = {{{GLM}}: {{General Language Model Pretraining}} with {{Autoregressive Blank Infilling}}},
  author = {Du, Zhengxiao and Qian, Yujie and Liu, Xiao and Ding, Ming and Qiu, Jiezhong and Yang, Zhilin and Tang, Jie},
  year = {2022},
  eprint = {2103.10360},
  archivePrefix={arXiv}
}

@misc{Gao_2024_Retrieval-Augmented,
  title = {Retrieval-{{Augmented Generation}} for {{Large Language Models}}: {{A Survey}}},
  author = {Gao, Yunfan and Xiong, Yun and Gao, Xinyu and Jia, Kangxiang and Pan, Jinliu and Bi, Yuxi and Dai, Yi and Sun, Jiawei and Wang, Meng and Wang, Haofen},
  year = {2024},
  eprint = {2312.10997},
  archivePrefix={arXiv}
}

@misc{Hambardzumyan_2022_Deepa,
  title = {Deep {{Lake}}: A {{Lakehouse}} for {{Deep Learning}}},
  author = {Hambardzumyan, Sasun and Tuli, Abhinav and Ghukasyan, Levon and Rahman, Fariz and Topchyan, Hrant and Isayan, David and McQuade, Mark and Harutyunyan, Mikayel and Hakobyan, Tatevik and Stranic, Ivo and Buniatyan, Davit},
  year = {2022},
  eprint = {2209.10785},
  archivePrefix={arXiv}
}

@article{Kasneci_2023_ChatGPT,
  title={ChatGPT for good? On opportunities and challenges of large language models for education},
  author={Kasneci, Enkelejda and Se{\ss}ler, Kathrin and K{\"u}chemann, Stefan and Bannert, Maria and Dementieva, Daryna and Fischer, Frank and Gasser, Urs and Groh, Georg and G{\"u}nnemann, Stephan and H{\"u}llermeier, Eyke and others},
  journal={Learning and individual differences},
  volume={103},
  pages={102274--102282},
  year={2023},
}

@misc{Liu_2024_PropertyGPT,
  title = {{{PropertyGPT}}: {{LLM-driven Formal Verification}} of {{Smart Contracts}} through {{Retrieval-Augmented Property Generation}}},
  author = {Liu, Ye and Xue, Yue and Wu, Daoyuan and Sun, Yuqiang and Li, Yi and Shi, Miaolei and Liu, Yang},
  year = {2024},
  eprint = {2405.02580},
  archivePrefix={arXiv}
}

@inproceedings{Napoli_2023_Evaluating,
  title={Evaluating ChatGPT for Smart Contracts Vulnerability Correction},
  author={Napoli, Emanuele Antonio and Gatteschi, Valentina},
  booktitle={Proceedings of the IEEE Annual Computers, Software, and Applications Conference (COMPSAC)},
  pages={1828--1833},
  year={2023}
}

@misc{Shou_2024_LLM4Fuzz,
  title = {{{LLM4Fuzz}}: {{Guided Fuzzing}} of {{Smart Contracts}} with {{Large Language Models}}},
  author = {Shou, Chaofan and Liu, Jing and Lu, Doudou and Sen, Koushik},
  year = {2024},
  eprint = {2401.11108},
  archivePrefix={arXiv}
}

@inproceedings{Storhaug_2023_Efficient,
  title = {Efficient {{Avoidance}} of {{Vulnerabilities}} in {{Auto-completed Smart Contract Code Using Vulnerability-constrained Decoding}}},
  booktitle = {Proceedings of the {{IEEE}} {{International Symposium}} on {{Software Reliability Engineering}} ({{ISSRE}})},
  author = {Storhaug, André and Li, Jingyue and Hu, Tianyuan},
  year = {2023},
  pages = {683--693}
}

@misc{Xia_2024_AuditGPT,
  title = {{{AuditGPT}}: {{Auditing Smart Contracts}} with {{ChatGPT}}},
  author = {Xia, Shihao and Shao, Shuai and He, Mengting and Yu, Tingting and Song, Linhai and Zhang, Yiying},
  year = {2024},
  eprint = {2404.04306},
  archivePrefix={arXiv}
}

@inproceedings{Yang_2024_Automated,
  title = {Automated {{Smart Contract Vulnerability Detection}} Using {{Fine-tuned Large Language Models}}},
  booktitle = {Proceedings of the {{International Conference}} on {{Blockchain Technology}} and {{Applications}} (ICBTA)},
  author = {Yang, Zhiju and Man, Gaoyuan and Yue, Songqing},
  year = {2024},
  pages = {19--23}
}

@misc{Zheng__TAKE,
      title={Take a Step Back: Evoking Reasoning via Abstraction in Large Language Models}, 
      author={Huaixiu Steven Zheng and Swaroop Mishra and Xinyun Chen and Heng-Tze Cheng and Ed H. Chi and Quoc V Le and Denny Zhou},
      year={2024},
      eprint={2310.06117},
      archivePrefix={arXiv}
}

@inproceedings{Chaliasos_2024_Smart,
  title = {Smart {{Contract}} and {{DeFi Security Tools}}: {{Do They Meet}} the {{Needs}} of {{Practitioners}}?},
  booktitle = {Proceedings of the {{IEEE}}/{{ACM}} {{International Conference}} on {{Software Engineering}} (ICSE)},
  author = {Chaliasos, Stefanos and Charalambous, Marcos Antonios and Zhou, Liyi and Galanopoulou, Rafaila and Gervais, Arthur and Mitropoulos, Dimitris and Livshits, Benjamin},
  year = {2024},
  pages = {1--13}
}

@article{Sharma_2023_review,
  title = {A Review of Smart Contract-Based Platforms, Applications, and Challenges},
  author = {Sharma, Pratima and Jindal, Rajni and Borah, Malaya Dutta},
  year = {2023},
  journal = {Cluster Computing},
  volume = {26},
  number = {1},
  pages = {395--421}
}

@article{zheng2023dappscan,
  title={Dappscan: building large-scale datasets for smart contract weaknesses in dapp projects},
  author={Zheng, Zibin and Su, Jianzhong and Chen, Jiachi and Lo, David and Zhong, Zhijie and Ye, Mingxi},
  journal={IEEE Transactions on Software Engineering},
  year={2024},
   volume={50},
  number={6},
  pages={1360-1373}
}

@Inproceedings{DurieuxEtAl2020ICSE,
  title={Empirical Review of Automated Analysis Tools on 47,587 {Ethereum} Smart Contracts},
  author={Durieux, Thomas and Ferreira, Jo{\~a}o F. and Abreu, Rui and Cruz, Pedro},
  booktitle={Proceedings of the ACM/IEEE 42nd International conference on software engineering (ICSE)},
  pages={530--541},
  year={2020}
}

@inproceedings{Luu_2016_Making,
  title = {Making {{Smart Contracts Smarter}}},
  booktitle = {Proceedings of the {{ACM SIGSAC Conference}} on {{Computer}} and {{Communications Security}} (CCS)},
  author = {Luu, Loi and Chu, Duc-Hiep and Olickel, Hrishi and Saxena, Prateek and Hobor, Aquinas},
  year = {2016},
  pages = {254--269}
}

@inproceedings{Torres_2021_ConFuzzius,
  title = {{{ConFuzzius}}: {{A Data Dependency-Aware Hybrid Fuzzer}} for {{Smart Contracts}}},
  booktitle = {Proceedings of the {{IEEE European Symposium}} on {{Security}} and {{Privacy}} (EuroS\&P)},
  author = {Torres, Christof Ferreira and Iannillo, Antonio Ken and Gervais, Arthur and State, Radu},
  year = {2021},
  pages = {103--119},
}

@misc{wagner2018eip,
  author={Wagner, G},
  year={2018},
  title={Eip-1470: Smart contract weakness classification (SWC)},
  howpublished = {\url{https://github.com/ethereum/EIPs}},
  note = {Accessed: 2024-12-24}
}

@misc{slowmist,
  author = {SlowMist Zone},
  year = {2024},
  title = {SlowMist Hacked},
  howpublished = {\url{https://hacked.slowmist.io}},
  note = {Accessed: 2024-12-24}
}

@misc{mythril,
  author = {Mythril},
  year = {2024},
  title = {Mythril},
  howpublished = {\url{https://mythril-classic.readthedocs.io/}},
  note = {Accessed: 2024-12-24}
}

@misc{VelosoConkas,
  author = {Veloso, Nuno},
  year = {2024},
  title = {Conkas: {{A Modular}} and {{Static Analysis Tool}} for {{Ethereum Bytecode}}},
  howpublished = {\url{https://github.com/nveloso/conkas/}},
  note = {Accessed: 2024-12-24}
}

@inproceedings{li2024cobra,
  title={COBRA: Interaction-Aware Bytecode-Level Vulnerability Detector for Smart Contracts},
  author={Li, Wenkai and Li, Xiaoqi and Li, Zongwei and Zhang, Yuqing},
  booktitle={Proceedings of the 39th IEEE/ACM International Conference on Automated Software Engineering (ASE)},
  pages={1358--1369},
  year={2024}
}

@inproceedings{kong2024characterizing,
  title={Characterizing the Solana NFT Ecosystem},
  author={Kong, Dechao and Li, Xiaoqi and Li, Wenkai},
  booktitle={Companion Proceedings of the ACM on Web Conference (WWW)},
  pages={766--769},
  year={2024}
}

@inproceedings{niu2024unveiling,
  title={Unveiling Wash Trading in Popular NFT Markets},
  author={Niu, Yuanzheng and Li, Xiaoqi and Peng, Hongli and Li, Wenkai},
  booktitle={Companion Proceedings of the ACM on Web Conference (WWW)},
  pages={730--733},
  year={2024}
}

@article{li2023overview,
  title={An overview of AI and blockchain integration for privacy-preserving},
  author={Li, Zongwei and Kong, Dechao and Niu, Yuanzheng and Peng, Hongli and Li, Xiaoqi and Li, Wenkai},
  journal={arXiv preprint arXiv:2305.03928},
  year={2023}
}

@inproceedings{li2024stateguard,
  title={StateGuard: Detecting State Derailment Defects in Decentralized Exchange Smart Contract},
  author={Li, Zongwei and Li, Wenkai and Li, Xiaoqi and Zhang, Yuqing},
  booktitle={Companion Proceedings of the ACM on Web Conference (WWW)},
  pages={810--813},
  year={2024}
}

@inproceedings{li2024defitail,
  title={DeFiTail: DeFi Protocol Inspection through Cross-Contract Execution Analysis},
  author={Li, Wenkai and Li, Xiaoqi and Zhang, Yuqing and Li, Zongwei},
  booktitle={Companion Proceedings of the ACM on Web Conference (WWW)},
  pages={786--789},
  year={2024}
}

@article{mao2024scla,
  title={SCLA: Automated Smart Contract Summarization via LLMs and Semantic Augmentation},
  author={Mao, Yingjie and Li, Xiaoqi and Li, Wenkai and Wang, Xin and Xie, Lei},
  journal={arXiv preprint arXiv:2402.04863},
  year={2024}
}

@inproceedings{li2020characterizing,
  title={Characterizing erasable accounts in ethereum},
  author={Li, Xiaoqi and Chen, Ting and Luo, Xiapu and Yu, Jiangshan},
 booktitle={Proceedings of the 23rd International Conference on Information Security (ISC)},
  pages={352--371},
  year={2020}
}

@article{li2024guardians,
  title={Guardians of the ledger: Protecting decentralized exchanges from state derailment defects},
  author={Li, Zongwei and Li, Wenkai and Li, Xiaoqi and Zhang, Yuqing},
  journal={IEEE Transactions on Reliability},
  year={2024}
}

@inproceedings{Zhuang_2020_Smart,
  title     = {Smart contract vulnerability detection using graph neural networks},
  author    = {Zhuang, Yuan and Liu, Zhenguang and Qian, Peng and Liu, Qi and Wang, Xiang and He, Qinming},
  booktitle = {Proceedings of the {Twenty-Ninth International Conference} on {International Joint Conferences on Artificial Intelligence} ({IJCAI})},
  pages     = {3283--3290},
  year      = {2021}
}

@inproceedings{Feist_2019_Slither,
  title     = {Slither: {{A Static Analysis Framework}} for {{Smart Contracts}}},
  booktitle = {Proceedings of the {{IEEE}}/{{ACM}} {{International Workshop}} on {{Emerging Trends}} in {{Software Engineering}} for {{Blockchain}} (WETSEB)},
  author    = {Feist, Josselin and Grieco, Gustavo and Groce, Alex},
  year      = {2019},
  pages     = {8--15}
}

@inproceedings{tsankov2018securify,
  title     = {Securify: Practical security analysis of smart contracts},
  author    = {Tsankov, Petar and Dan, Andrei and Drachsler-Cohen, Dana and Gervais, Arthur and Buenzli, Florian and Vechev, Martin},
  booktitle = {Proceedings of the {ACM SIGSAC Conference} on {Computer and Communications Security} (CCS)},
  pages     = {67--82},
  year      = {2018}
}

@inproceedings{brent2020ethainter,
  title     = {Ethainter: a smart contract security analyzer for composite vulnerabilities},
  author    = {Brent, Lexi and Grech, Neville and Lagouvardos, Sifis and Scholz, Bernhard and Smaragdakis, Yannis},
  booktitle = {Proceedings of the ACM SIGPLAN Conference on Programming Language Design and Implementation (PLDI)},
  pages     = {454--469},
  year      = {2020}
}

@article{brent2018vandal,
  title   = {Vandal: A scalable security analysis framework for smart contracts},
  author  = {Brent, Lexi and Jurisevic, Anton and Kong, Michael and Liu, Eric and Gauthier, Francois and Gramoli, Vincent and Holz, Ralph and Scholz, Bernhard},
  journal = {arXiv preprint arXiv:1809.03981},
  year    = {2018}
}

@inproceedings{grieco2020echidna,
  title     = {Echidna: effective, usable, and fast fuzzing for smart contracts},
  author    = {Grieco, Gustavo and Song, Will and Cygan, Artur and Feist, Josselin and Groce, Alex},
  booktitle = {Proceedings of the ACM SIGSOFT international symposium on software testing and analysis (ISSTA)},
  pages     = {557--560},
  year      = {2020}
}

@inproceedings{wustholz2020harvey,
  title     = {Harvey: A greybox fuzzer for smart contracts},
  author    = {W{\"u}stholz, Valentin and Christakis, Maria},
  booktitle = {Proceedings of the ACM Joint Meeting on European Software Engineering Conference and Symposium on the Foundations of Software Engineering (FSE)},
  pages     = {1398--1409},
  year      = {2020}
}

@article{mueller2018smashing,
  title   = {Smashing ethereum smart contracts for fun and real profit},
  author  = {Mueller, Bernhard},
  journal = {HITB SECCONF Amsterdam},
  volume  = {9},
  number  = {54},
  pages   = {4--17},
  year    = {2018}
}

@inproceedings{Mossberg_2019_Manticore,
  title     = {Manticore: {{A User-Friendly Symbolic Execution Framework}} for {{Binaries}} and {{Smart Contracts}}},
  booktitle = {Proceedings of the {34th {{IEEE}}/{{ACM International Conference}} on {{Automated Software Engineering}}} (ASE)},
  author    = {Mossberg, Mark and Manzano, Felipe and Hennenfent, Eric and Groce, Alex and Grieco, Gustavo and Feist, Josselin and et al.},
  year      = {2019},
  pages     = {1186--1189}
}

@inproceedings{luo2024scvhunter,
  title     = {Scvhunter: Smart contract vulnerability detection based on heterogeneous graph attention network},
  author    = {Luo, Feng and Luo, Ruijie and Chen, Ting and Qiao, Ao and He, Zheyuan and Song, Shuwei and Jiang, Yu and Li, Sixing},
  booktitle = {Proceedings of the IEEE/ACM 46th international conference on software engineering (ICSE)},
  pages     = {1--13},
  year      = {2024}
}

@inproceedings{schneidewind2020ethor,
  title     = {ethor: Practical and provably sound static analysis of ethereum smart contracts},
  author    = {Schneidewind, Clara and Grishchenko, Ilya and Scherer, Markus and Maffei, Matteo},
  booktitle = {Proceedings of the ACM SIGSAC Conference on Computer and Communications Security (CCS)},
  pages     = {621--640},
  year      = {2020}
}

@inproceedings{torres2018osiris,
  title     = {Osiris: Hunting for integer bugs in ethereum smart contracts},
  author    = {Torres, Christof Ferreira and Sch{\"u}tte, Julian and State, Radu},
  booktitle = {Proceedings of the 34th annual computer security applications conference (ACSAC)},
  pages     = {664--676},
  year      = {2018}
}

@misc{pakala,
  title  = {Offensive vulnerability scanner for ethereum, and symbolic execution tool for the Ethereum Virtual Machine},
  author = {Palkeo},
  note   = {\url{https://github.com/palkeo/pakala/}},
  year   = {2025}
}

@article{chen2025chatgpt,
  title   = {When chatgpt meets smart contract vulnerability detection: How far are we?},
  author  = {Chen, Chong and Su, Jianzhong and Chen, Jiachi and Wang, Yanlin and Bi, Tingting and Yu, Jianxing and Wang, Yanli and Lin, Xingwei and Chen, Ting and Zheng, Zibin},
  journal = {ACM Transactions on Software Engineering and Methodology (TOSEM)},
  volume  = {34},
  number  = {4},
  pages   = {1--30},
  year    = {2025}
}

@inproceedings{chen2024improving,
  title     = {Improving smart contract security with contrastive learning-based vulnerability detection},
  author    = {Chen, Yizhou and Sun, Zeyu and Gong, Zhihao and Hao, Dan},
  booktitle = {Proceedings of the IEEE/ACM 46th International Conference on Software Engineering (ICSE)},
  pages     = {1--11},
  year      = {2024}
}

@article{liu2025detecting,
  title   = {Detecting Smart Contract State-Inconsistency Bugs via Flow Divergence and Multiplex Symbolic Execution},
  author  = {Liu, Yinxi and Meng, Wei and Zhang, Yinqian},
  journal = {Proceedings of the ACM on Software Engineering},
  volume  = {2},
  number  = {FSE},
  pages   = {22--43},
  year    = {2025}
}

@inproceedings{song2025silence,
  title     = {Silence False Alarms: Identifying Anti-Reentrancy Patterns on Ethereum to Refine Smart Contract Reentrancy Detection},
  author    = {Song, Qiyang and Huang, Heqing and Jia, Xiaoqi and Xie, Yuanbo and Cao, Jiahao},
  booktitle = {Proceedings of the Network and Distributed System Security (NDSS)},
  year      = {2025},
  pages     = {1--18}
}

@article{shang2025cegt,
  title   = {CEGT: Smart contract vulnerability detection via Connectivity-Enhanced GCN-Transformer},
  author  = {Shang, Jiandong and Li, Jiaru and Sui, Yizhe and Guo, Hengliang and Gao, Xu and Zhang, Dujuan and Guo, Yang and Wu, Gang},
  journal = {Journal of Systems and Software},
  pages   = {112454--112465},
  year    = {2025}
}

@article{zhen2024gnn,
  title   = {DA-GNN: A smart contract vulnerability detection method based on Dual Attention Graph Neural Network},
  author  = {Zhen, Zixian and Zhao, Xiangfu and Zhang, Jinkai and Wang, Yichen and Chen, Haiyue},
  journal = {Computer Networks},
  volume  = {242},
  pages   = {110238--110248},
  year    = {2024}
}

@inproceedings{10.1145/3543507.3583367,
  author    = {Qian, Peng and Liu, Zhenguang and Yin, Yifang and He, Qinming},
  title     = {Cross-Modality Mutual Learning for Enhancing Smart Contract Vulnerability Detection on Bytecode},
  year      = {2023},
  booktitle = {Proceedings of the ACM Web Conference (WWW)},
  pages     = {2220--2229}
}

@article{ma2025uncovering,
  author  = {Ma, Zuchao and Jiang, Muhui and Luo, Xiapu and Wang, Haoyu and Zhou, Yajin},
  journal = {IEEE Transactions on Dependable and Secure Computing},
  title   = {Uncovering NFT Domain-Specific Defects on Smart Contract Bytecode},
  year    = {2025},
  volume  = {22},
  number  = {5},
  pages   = {4877-4895}
}

@inproceedings{wang2023smart,
  author    = {Wang, Yichuan and Zhao, Jingjing and Zhang, Yaling and Hei, Xinhong and Zhu, Lei},
  title     = {Smart contract symbol execution vulnerability detection method based on {CFG} path pruning},
  booktitle = {Proceedings of the 5th {ACM} International Symposium on Blockchain and Secure Critical Infrastructure},
  year      = {2023},
  pages     = {132--139}
}

@article{10700860,
  author  = {Wang, Yichen and Zhao, Xiangfu and He, Long and Zhen, Zixian and Chen, Haiyue},
  journal = {IEEE Transactions on Network Science and Engineering},
  title   = {ContractGNN: Ethereum Smart Contract Vulnerability Detection Based on Vulnerability Sub-Graphs and Graph Neural Networks},
  year    = {2024},
  volume  = {11},
  number  = {6},
  pages   = {6382-6395}
}

@article{10.1145/3699597,
  author  = {Xiang, Jianhang and Gao, Zhipeng and Bao, Lingfeng and Hu, Xing and Chen, Jiayuan and Xia, Xin},
  title   = {Automating Comment Generation for Smart Contract from Bytecode},
  year    = {2025},
  volume  = {34},
  number  = {3},
  journal = {ACM Transactions on Software Engineering and Methodology},
  pages   = {1-31}
}

@article{10486822,
  author  = {Zheng, Zibin and Su, Jianzhong and Chen, Jiachi and Lo, David and Zhong, Zhijie and Ye, Mingxi},
  journal = {IEEE Transactions on Software Engineering},
  title   = {DAppSCAN: Building Large-Scale Datasets for Smart Contract Weaknesses in DApp Projects},
  year    = {2024},
  volume  = {50},
  number  = {6},
  pages   = {1360-1373}
}

@inproceedings{10646885,
  author    = {Wang, Sally Junsong and Pei, Kexin and Yang, Junfeng},
  booktitle = {Proceedings of the 2024 IEEE Symposium on Security and Privacy (SP)},
  title     = {SmartInv: Multimodal Learning for Smart Contract Invariant Inference},
  year      = {2024},
  pages     = {2217-2235}
}

@article{LI2026122645,
  title   = {An attack detection mechanism in smart contracts based on deep learning and feature fusion},
  journal = {Information Sciences},
  volume  = {722},
  pages   = {122645--122664},
  year    = {2025},
  author  = {Peiqiang Li and Guojun Wang and Wanyi Gu and Xubin Li and Xiangyong Liu and Yuheng Zhang}
}

@misc{runtimeverification,
  author       = {{Runtime Verification Inc.}},
  title        = {Runtime Verification: Formal Verification for Blockchain Security},
  year         = {2026},
  howpublished = {\url{https://runtimeverification.com/}}
}

\end{document}